\documentclass[conference]{IEEEtran}
\ifCLASSINFOpdf
\else
\fi
\usepackage{graphicx}
\usepackage{amsmath}
\usepackage{lineno}
\usepackage{amssymb}
\usepackage[linesnumbered,ruled,vlined]{algorithm2e}
\usepackage{enumerate}
\usepackage{syntonly}
\usepackage{multicol}
\usepackage{array}
\usepackage[footnotesize]{subfigure}
\usepackage{stfloats}
\usepackage{url}
\usepackage{threeparttable}
\usepackage{multirow}
\usepackage{tabularx}

\usepackage{url}
\usepackage{amsmath}
\usepackage{amssymb}
\usepackage[linesnumbered,ruled,vlined]{algorithm2e}
\usepackage{enumerate}
\usepackage{syntonly}
\usepackage{multicol}
\usepackage{array}
\usepackage[footnotesize]{subfigure}
\usepackage{stfloats}
\usepackage{url}
\usepackage{threeparttable}
\usepackage{cite}
\usepackage{multirow}
\usepackage{color}

\usepackage{fancyhdr}
\SetAlFnt{\small}
\SetAlCapFnt{\small}
\SetAlCapNameFnt{\small}
\SetAlCapHSkip{0pt}
\IncMargin{-\parindent}

\makeatletter
\def\firstfoot{\def\@firstfoot{}}
\def\runningfoot{\def\@runningfoot{}}

\makeatother

\usepackage{soul}

\begin{document}


\title{Source Camera Identification and Detection in Digital Videos through Blind Forensics }

\author{\IEEEauthorblockN{Venkata Udaya Sameer, Shilpa Mukhopadhyay, Ruchira Naskar\\ and Ishaan Dali}
\IEEEauthorblockA{Department of Computer Science and Engineering\\
National Institute of Technology, Rourkela\\
 India--769008\\
Email: \{515cs1003,113cs0156,naskar,713cs2136\}@nitrkl.ac.in.}
}

\maketitle

\begin{abstract}

\emph{Source camera identification} in digital videos is the problem of associating an unknown digital video with its source device, within a closed set of possible devices. 
The existing techniques in source detection of digital videos try to find a fingerprint of the actual source in the video in form of PRNU (Photo Response Non--Uniformity), and match it against the SPN (Sensor Pattern Noise) of each possible device. The highest correlation indicates the correct source. We investigate the problem of identifying a video source through a feature based approach using machine learning.  
In this paper, we present a blind forensic technique of video source  authentication and identification, based on feature extraction, feature selection and subsequent source classification. 
The main aim is to determine whether a claimed source for a video is actually its original source. If not, we identify its original source. 
Our experimental results prove the efficiency of the proposed method compared to traditional fingerprint based technique. 

\end{abstract}



\section{Introduction}
\label{sec:introduction}

\emph{Digital Forensics} refers to the branch of science dealing with examination and recovery of evidences that are left behind in electronic devices with digital storage, such as cell phones, computers and digital cameras. 
It is more challenging compared to traditional forensic science, because of the fact that crimes committed using digital devices leave behind minimal traces. Digital Forensics includes the recovery and investigation of clues/traces from digital devices, associated to cyber crimes. One of the main motives of carrying out investigation and analysis regarding authenticity of digital media or their sources, is to establish those as evidences in the court of law, against cyber criminals. Multimedia data including text, audio, images or videos often act as the primary sources of digital evidences in legal cases, or media/broadcast industries. In the paper, we deal with blind source authentication of digital videos.

In this paper, our focus is on \emph{video forensics}, where digital videos are considered to be the main digital evidence in forensic analysis. 
\footnote{This work is funded by Board of Research in Nuclear Sciences (BRNS), Department of Atomic Energy (DAE), Govt. of India, Grant No. 34/20/22/2016-BRNS/34363, dated: 16/11/2016.}
Videos often act as major sources of evidence towards proving the credibility of events in legal controversies. 
In such scenarios, it gains extreme importance to authenticate videos, as well as their sources, and hence establish their reliability and trustworthiness.
However, it is of paramount importance to verify that the data is not tampered with previously, before produce it as a legal evidence in the court of law.
Additionally, the rapid emergence of a wide range of cheap, easy-to-use video-editing tools today, such as Adobe After-effects, Adobe Premiere-pro etc., have rendered the possibility of easy modification and tampering of data in digital videos, much higher than before.

Along with data authentication, data-origin authentication of digital videos is equally important. 
The main question regarding this issue is: Did the video indeed originate from the device that it is claimed to be? Or whether the claim is false? In real life, validating the answer to the above, find application in movie piracy detection. 
In many cases, source authentication plays an important role to disprove false accusation of an innocent, and subsequently find the actual criminal.

In this paper, we try to provide an authoritative answer to the above question. 
Our major aim is to map a digital video to its source camera, in a completely \emph{blind} way, that is without making any assumption of a--priori information processing and storage (such as embedding watermarks or fingerprints into the videos). 
This makes the proposed approach completely independent of data pre-processing, and assists us in 100\% post-processing based forensic source identification. 
Moreover, it does not require any specialized software or hardware chips, (such as those used in watermarking or fingerprinting of data), hence bringing down the cost of such devices.

In this paper, we provide the video source identification problem as a machine learning based classification problem. The aim is to predict the source for each video. The possible sources form a closed set, each of which corresponds to one class. The classifiers are trained with appropriate features extracted from the training set of video frames.
Our major contributions in this paper include the followings. 
(1) Identifying efficient features for video source identification, as a classification problem.
(2) To investigate and analyze the accuracy of source identification by the model, based on training using the above identified features. 
This would lead to the development of an efficient video source authentication and detection model through blind forensics. 

Rest of the paper is organized as follows. 
In Section~\ref{sec:related_work}, we discuss the related researches in the field of video forensics and source camera identification. 
In Section~\ref{sec:Source Identification}, we present a detailed description of the proposed video source classification method, including detailed feature extraction and selection procedures. In Section~\ref{sec:results_and_discussion}, we present our experimental results and discussion. 
In Section~\ref{sec:conclusion}, we conclude the paper and give directions for future work in this area.

\section{Background}
\label{sec:related_work}


A survey of video forensics \cite{milani2012overview} was done in 2012, in which the authors studied various existing forensic tools for acquisition, compression and modification detection of digital videos. 
Another overview on detection of tampering in digital videos \cite{sitara2016digital} was published in 2016, where the authors explored passive video forgery detection based on three kinds of forgeries, viz., detecting double or multiple compression, detecting region tampering and detecting video inter-frame forgery. 
They surveyed a number of works done related to those three kinds of forgeries individually. 
A survey on techniques regarding video content authentication \cite{singh2017video} has been performed in 2017. Here, the authors investigated intra-frame as well as inter-frame forgery detection through source camera identification, tampering detection and hidden data recovery.

In the domain of source camera identification, one of the pioneer works was that of Kharrazi et al. \cite{kharrazi2004blind}. Kharrazi et al. \cite{kharrazi2004blind} proposed a source camera authentication scheme for digital images through blind digital forensics. 
They extracted features according to the belief that output image in a digital camera is immensely affected by CFA configuration and democaicing algorithm and by color processing or transformation. Based on this concept, they extracted features and classified using SVM classifier. Kurosawa et al. \cite{kurosawa1999ccd} proposed a CCD based fingerprint to map video camera from video taped images. It is based on the primary concept of non uniformity of black currents on CCD chips because of different dark current generations of certain pixels than others. Since the location of the pixels are fixed, a fixed pattern noise is generated and this fact is exploited in their experiment. Lucas et al.~\cite{lukas2006digital} proposed a method for image source detection using sensor pattern noise as the unique fingerprint for each camera. However, it is an active forensic method as the watermark used is reference pattern noise to map an image to its camera. The effectiveness of camera identification \cite{goljan2016effect} was studied based on the effects of compression on images and videos. Their work explored the change in detection thresholds due to lossy compression. Also, Chen et.al.~\cite{chen2015live} propose a fingerprint based technique for video source identification in lossy wireless frameworks.

\section{Source Camera Identification Model}
\label{sec:Source Identification}

In the recent years, the research interest in the problem of source identification has gained rapid growth. 
In this paper, we deal with source identification of digital videos, based on machine learning classification. 
In the subsequent sections we present the detailed feature extraction procedure along with description of all the features used by our video source classification model. 
The detailed results are presented in Section~\ref{sec:results_and_discussion}. 

A number of researchers have targeted source identification through digital forensics in the recent years, majority of those addressing image source identification. 
In this work, the set of features adopted by us for video source identification, are based on those used by Kharrazi et al. in \cite{kharrazi2004blind}, for image source classification. 

\subsection{Feature Extraction}
\label{sec:Classification of Image Sources}

In out proposed model, we use three sets of video features for feature extraction, they are described as follows.

\begin{itemize}
\item [a.] \textbf{Image Quality Metrics(IQM)}\\
IQM extracts 40 features according to \cite{sayood2002statistical} and \cite{avcibas2003steganalysis} which are applied by comparing to the following four deviations from original image: addition of noise, filtering with Gaussian filter, JPEG compression and SPIHT compression. 10 features for each of 4 distorted versions of images are computed, giving a total of 40 features. The 10 features extracted for each of the above four are listed below:

Assumptions: $C_k(i,j)$ represents the multispectral component of an image at pixel position $(i,j)$ and band $k$. $C(i,j)$ represents the multispectral pixel vectors at position $(i,j)$. $\hat{C}$ represents the distorted version of an image. $M_i$ represents the $i^{th}$ IQM measure. $R, C, K$ represent the number of rows, column and bands in an image respectively.

\begin{itemize}
\item [i.] \textbf{Minkowsky Measures}
\begin{itemize}
\item [1.] \textbf{Mean Absolute Error(MAE)}\\
$M_1=\frac{1}{K}\sum_{k=1}^{K}\frac{1}{R\times C}{\sum_{i=1}^{R}\sum_{j=1}^{C}|C_{k}(i,j)-\hat{C_{k}}(i,j)|}$\\
\item [2.] \textbf{Mean Square Error(MSE)}\\
$M_2=\frac{1}{K}\sum_{k=1}^{K}\sqrt{\frac{1}{R\times C}{\sum_{i=1}^{R}\sum_{j=1}^{C}|C_{k}(i,j)-\hat{C_{k}}(i,j)|^2}}$\\
\end{itemize}
\item [ii.] \textbf{Correlation-based Measures}
\begin{itemize}
\item [1.] \textbf{Czekanowski Distance}\\
$M_3=\frac{1}{RC}{\sum_{i=1}^{R}\sum_{j=1}^{C} 1- \frac{2\times \sum_{k=1}^{K}min(C_{k}(i,j),\hat{C_{k}}(i,j))}{\sum_{k=1}^{K}(C_{k}(i,j)+\hat{C_{k}}(i,j))}}$\\
\item [2.] \textbf{Cross-Correlation Measure}\\
$M_4=1-\frac{1}{K}\sum_{k=1}^{K}\frac{\sum_{i=1}^{R}\sum_{j=1}^{C}[C_{k}(i,j)-\hat{C_{k}}(i,j)]^2}{\sum_{i=1}^{R}\sum_{j=1}^{C}[C_{k}(i,j)]^2}$\\
\item [3.] \textbf{Normalized Cross-Correlation Measure}\\
$M_5=\frac{1}{K}\sum_{k=1}^{K}\frac{\sum_{i=1}^{R}\sum_{j=1}^{C} C_{k}(i,j)\cdot\hat{C_{k}}(i,j)}{\sum_{i=1}^{R}\sum_{j=1}^{C}[C_{k}(i,j)]^2}$\\
\item [4.] \textbf{Statistics of angles between pixel vectors of 2 images}\\
$M_6=1-{\frac{1}{R\times C}\sum_{i=1}^{R}\sum_{j=1}^{C} \frac{2}{\pi}\arccos \frac{<C(i,j),\hat{C}(i,j)>}{\| C(i,j) \| \| \hat{C}(i,j) \|}}$\\
\end{itemize}
\item [iii.] \textbf{Spectral Measures}
Let DFT of k-th band of original and distorted image be $M_k(u,v)$ and $\hat{M_k}(u,v)$ respectively.
Phase=$P(u,v)$=$\arctan M(u,v)$
Magnitude=$|M(u,v)|$
\begin{itemize}
\item [1.] \textbf{Spectral Magnitude}\\
$M_7=\frac{1}{K \times R \times C}\sum_{k=1}^{K}\sum_{i=1}^{R}\sum_{j=1}^{C}||M(u,v)|-|\hat{M}(u,v)||^2$\\
\item [2.] \textbf{Spectral Phase Distortion}\\
$M_8=\frac{1}{K \times R \times C}\sum_{k=1}^{K}\sum_{i=1}^{R}\sum_{j=1}^{C}||P(u,v)|-|\hat{P}(u,v)||^2$\\
\end{itemize}
\item [iv.] \textbf{Human Visual System (HVS) based Measures }
The HVS is assumed to be a bandpass filter as:\\
\[ H(\rho) =
  \begin{cases}
    0.05 e^{\rho^{0.554}} & \quad \text{if }\rho < 7\\
    e^{-9[|\log_{10} \rho - \log_{10} 9|]^{2.3}}  & \quad \text{if } \rho \geq 7\\
  \end{cases}
\]
where $\rho = \sqrt{u^2 + v^2}$ and ($u$, $v$) are pixel coordinates.
\begin{itemize}
\item [1.] \textbf{HVS measure}\\
$M_9=\frac{1}{K}\sum_{k=1}^{K} \frac{\sum_{i=1}^{R}\sum_{j=1}^{C}(U[C_k(i,j)]-U[\hat{C_k}(i,j)])}{\sum_{i=1}^{R}\sum_{j=1}^{C}(U[C_k(i,j)])^2}$ \\\\
where $U[C_k(i,j)]=Inverse DCT\{H(\rho) \times DCT(u,v)\}$
\item [2.] \textbf{Laplacian Mean Square Error(LMSE)}\\\\
$M_{10}=\frac{\sum_{i=1}^{R-1}\sum_{j=2}^{C-1}[O\{F(i,j)\}-O\{\hat{F}(i,j)\}]^2}{\sum_{i=1}^{R-1}\sum_{j=2}^{C-1}[O\{F(i,j)\}]^2} $ \\\\
where $O\{F(i,j)\}=F(i+1,j)+F(i-1,j)+F(i,j+1)+F(i,j-1)-4 F(i,j)$

\end{itemize}
\end{itemize}

\item [b.] \textbf{Non-Image Quality Metrics(Non-IQM)}\\
A total of 12 features are extracted by Non-IQM. The features extracted by Non-IQM are listed below.\\
Assumptions:
The R, G, B components of an RGB image are considered as $R(i,j)$, $G(i,j)$ and $B(i,j)$ respectively. K is the number of color bands (here K=3 for R,G,B).
\begin{itemize}
\item [i.] \textbf{Average Pixel Value Mean}\\
$\overline{R}=\frac{1}{R \times C} \sum_{i=1}^{R}\sum_{j=1}^{C} R(i,j)$\\
$\overline{G}=\frac{1}{R \times C} \sum_{i=1}^{R}\sum_{j=1}^{C} G(i,j)$\\
$\overline{B}=\frac{1}{R \times C} \sum_{i=1}^{R}\sum_{j=1}^{C} B(i,j)$\\
\item [ii.] \textbf{RGB Pairwise Correlation}\\\\
$RG=\frac{\sum_{i=1}^{R}\sum_{j=1}^{C} (R(i,j)-\overline{R})(G(i,j)-\overline{G})}{\sqrt{\sum_{i=1}^{R}\sum_{j=1}^{C} (R(i,j)-\overline{R})^2} \sqrt{\sum_{i=1}^{R}\sum_{j=1}^{C} (G(i,j)-\overline{G})^2}} $\\\\
$GB=\frac{\sum_{i=1}^{R}\sum_{j=1}^{C} (G(i,j)-\overline{G})(B(i,j)-\overline{B})}{\sqrt{\sum_{i=1}^{R}\sum_{j=1}^{C} (G(i,j)-\overline{G})^2} \sqrt{\sum_{i=1}^{R}\sum_{j=1}^{C} (B(i,j)-\overline{B})^2}} $\\\\
$BR=\frac{\sum_{i=1}^{R}\sum_{j=1}^{C} (B(i,j)-\overline{B})(R(i,j)-\overline{R})}{\sqrt{\sum_{i=1}^{R}\sum_{j=1}^{C} (B(i,j)-\overline{B})^2} \sqrt{\sum_{i=1}^{R}\sum_{j=1}^{C} (R(i,j)-\overline{R})^2}} $\\
\item [iii.] \textbf{RGB Pairwise Energy Ratio}\\
$E_1=\frac{E_G}{E_B}$\\
$E_2=\frac{E_G}{E_R}$\\
$E_3=\frac{E_B}{E_R}$\\
where:\\
$E_R=\sum_{i=1}^{R}\sum_{j=1}^{C} R(i,j)$\\
$E_G=\sum_{i=1}^{R}\sum_{j=1}^{C} G(i,j)$\\
$E_B=\sum_{i=1}^{R}\sum_{j=1}^{C} B(i,j)$\\
\item [iv.] \textbf{Neighbour Distribution Centre of Mass}\\
A similar image captured by 2 different source camera same similar distribution, but one is shifted from the other. Taking advantage of this fact, the shift is considered to be the centre of mass of neighbourhood plot. It is calculated in the following way for each of R, G, B components of the image separately:\\
$COM_{R}=hg_{1R}-hg_{2R}$\\
$COM_{G}=hg_{1G}-hg_{2G}$\\
$COM_{B}=hg_{1B}-hg_{2B}$\\\\
where:\\
$hg_{1R}=sum((hg_R(2:254)-hg_R(1:253))+(hg_R(2:254)+hg_R(3:255)))$\\
$hg_{2R}=sum((hg_R(3:255)-hg_R(2:254))+(hg_R(3:255)+hg_R(4:256)))$\\\\
$hg_{1G}=sum((hg_G(2:254)-hg_G(1:253))+(hg_G(2:254)+hg_G(3:255)))$\\
$hg_{2G}=sum((hg_G(3:255)-hg_G(2:254))+(hg_G(3:255)+hg_G(4:256)))$\\\\
$hg_{1B}=sum((hg_B(2:254)-hg_B(1:253))+(hg_B(2:254)+hg_B(3:255)))$\\
$hg_{2B}=sum((hg_B(3:255)-hg_B(2:254))+(hg_B(3:255)+hg_B(4:256)))$\\\\

Here $hg_R=histogram(R)$, $hg_G=histogram(G)$ and $hg_\textbf{}=histogram(B)$.

\end{itemize}
\item [c.] \textbf{Higher Order Wavelet Statistics (HOWS)}\\
HOWS are 36 features according to \cite{farid2003higher}. The algorithm decomposes the image into 9 smaller sub-images(level-3 according to Haar Wavelet Decomposition). For each of the 9 sub-images, 4 features are computed, giving a total of 36 features. The features extracted by HOWS are listed below.
\begin{itemize}
\item [i.] \textbf{Mean}\\
$\overline{X}=\frac{1}{R \times C} \sum_{i=1}^{R}\sum_{j=1}^{C} Img(i,j)$
\item [ii.] \textbf{Variance}\\
$Var=\frac{1}{R \times C-1} \sum_{i=1}^{R}\sum_{j=1}^{C} [Img(i,j)-\overline{X}]^2$
\item [iii.] \textbf{Skewness}\\
$Skewness=\frac{1}{R \times C \times \sigma ^ {3}} \sum_{i=1}^{R}\sum_{j=1}^{C} [Img(i,j)-\overline{X}]^3$
\item [iv.] \textbf{Kurtosis}\\
$Kurtosis=\frac{1}{R \times C \times \sigma ^ {4}} \sum_{i=1}^{R}\sum_{j=1}^{C} [Img(i,j)-\overline{X}]^4 - 3$\\\\
where $\sigma= \sqrt{Var}$ i.e.$\sigma$ is the Standard Deviation.
\end{itemize}
\end{itemize}

Hence total number of features extracted for each frame= 40(IQM)+12(Non-IQM)+36(HOWS)=88 features.\\

\subsection{Video Source Classification}
\label{sec:Source_Classification}
\begin{figure*}[t]
\centering
\includegraphics[clip,scale=0.11]{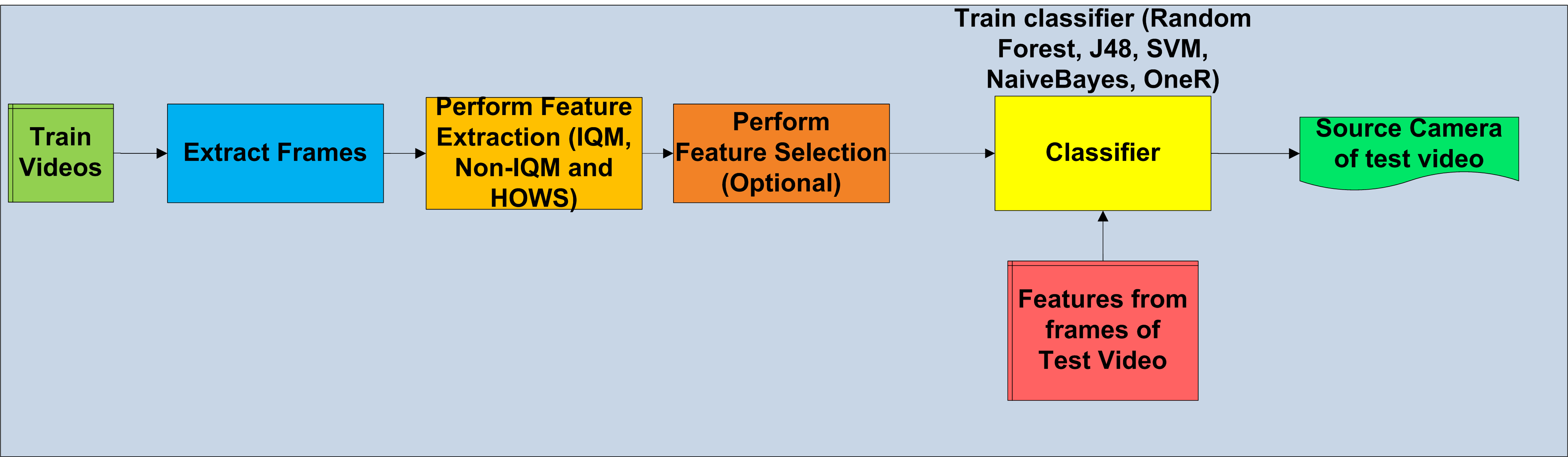}
\caption{Proposed video source classification model.}
\label{fig:model}
\end{figure*}

The proposed methodology for video source classification is depicted in form of a block diagram in Fig.~\ref{fig:model}. 
First, individual frames are extracted from the training videos, and then the features discussed in Section~\ref{sec:Classification of Image Sources} are extracted from each individual frame. 
A machine learning classifier is built, which is subsequently trained  based on the extracted feature matrix. 
After training with a sufficiently large video data set (training set), the classifier is ready to predict the source of an unknown video.
For a test video too, the same set of features are extracted from its constituent frames. 
The test feature set is used by the classifier to predict the source camera model of the test video. 
In our experiments we performed classification with multiple classifiers including Support Vector Machine (SVM), Random Forest, Tree based classifier in J48, Naive Bayes and a OneR classifier. 
The experiments are conducted in two phases, one in which feature selection is not performed and then using feature selection to reduce the feature dimensions, which eventually improved the performance of the proposed method. 
Attribute selectors used are Correlation Attribute evaluator~\cite{witten2016data} and OneR Attribute evaluator~\cite{witten2016data}, from which we obtain two different sets of top 30 features. 
In the proposed model, we consider the intersection of those two sets of 30 features each. 
That is, the common features belonging to both the sets of top 30 features, generated by the two attribute selectors, are considered in our work. 
The final features selected are 18 in number, from a total of 88 features used initially. 
We also use a 10--fold cross validation strategy in order to ensure that there is no overfitting in the proposed model.

%

\section{Experimental Results and Discussion}
\label{sec:results_and_discussion}

\begin{table}[t!]
\centering
\caption{Dataset for Source Detection
\label{tab:dataset}}{
\resizebox{\columnwidth}{!}{%
\begin{tabular}{l l l l l l}
\hline\hline
Camera & Resolution & \#Training Videos & \#Test Videos & \#Training Frames & \#Test Frames \\ [0.4ex] 
\hline 
Camera-1 & 480x848 & 70 & 53 & 14433 & 13991 \\ 
Camera-2 & 720x1280 & 70 & 58 & 13026 & 12603\\
Camera-3 & 720x1280 & 70 & 54 & 13264 & 10575\\
Camera-4 & 720x1280 & 70 & 56 & 13226 & 12309\\
Camera-5 & 720x1280 & 70 & 57 & 11010 & 10085\\ [1ex] 
\hline
\end{tabular}
}}
\end{table}

The details of cameras used in our experiments are given in Table~\ref{tab:dataset}. All the videos taken are of varied lengths (5-30 seconds), each 25--30 frames per second.%
Camera 1 through 5 in Table~\ref{tab:dataset}, are of following make and models: (Camera 1) Microsoft Lumia 540, (Camera 2) Samsung J7 Prime, (Camera 3) Moto E4, (Camera 4) Redmi 2 and (Camera 5) Canon Ixus 70.

For all the videos, we retrieve individual frames and extract the features discussed in Section~\ref{sec:Classification of Image Sources}. Then multiple classification algorithms are tested to find the optimal classifier. Our experimental results are shown in two phases. One, without using feature selection and the other, using the feature selection strategies. Table~\ref{table:without_feature_selection_result} gives the results of classification without using any feature selection. Table~\ref{table:feature_selection_result} gives the results of classification using feature selection by two attribute selectors (Correlation Attribute Evaluator and OneR Attribute Evaluator), and taking the common features out of the top 30, returned by both. 
A total of 18 features are selected.

\begin{table}[t!]
\centering
\caption{Experimental results without Feature Selection
\label{table:without_feature_selection_result}}{
\resizebox{\columnwidth}{!}{%
\scalebox{0.5}{
\begin{tabular}{c c c c c c} 
\hline\hline 
Algorithm & Camera & Precision & Recall & F-Measure & Overall Accuracy \\ [0.4ex] 
\hline 
 & Camera-1 & 0.999 & 1 & 0.999 &  \\      
 & Camera-2 & 0.997 & 0.996 & 0.997 & \\
Random Forest & Camera-3 & 0.997 & 0.999 & 0.998 & 0.9985\\
 & Camera-4 & 1 & 0.998 & 0.999 & \\
 & Camera-5 & 0.999 & 0.999 & 0.999 & \\ 
 & & & & & \\
  & Camera-1 & 0.997 & 0.997 & 0.997 &  \\   
 & Camera-2 & 0.983 & 0.985 & 0.994 & \\
J48 & Camera-3 & 0.993 & 0.993 & 0.993 & 0.9924\\
 & Camera-4 & 0.997 & 0.995 & 0.996 & \\
 & Camera-5 & 0.991 & 0.990 & 0.991 & \\ 
 & & & & & \\

  & Camera-1 & 0.897 & 0.9343 & 0.9152 &  \\   
 & Camera-2 & 0.993 & 0.8773 & 0.9315 & \\
SVM & Camera-3 & 0.972 & 0.9373 & 0.9543 & 0.891\\
 & Camera-4 & 0.867 & 0.8373 & 0.8518 & \\
 & Camera-5 & 0.838 & 0.9393 & 0.8857 & \\ 
 & & & & & \\

  & Camera-1 & 0.956 & 0.921 & 0.938 &  \\   
 & Camera-2 & 0.687 & 0.790 & 0.735 & \\
Naive Bayes & Camera-3 & 0.859 & 0.890 & 0.874 & 0.8449\\
 & Camera-4 & 0.877 & 0.741 & 0.803 & \\
 & Camera-5 & 0.870 & 0.880 & 0.875 & \\ 
 & & & & & \\
  & Camera-1 & 0.722 & 0.693 & 0.707 &  \\   
 & Camera-2 & 0.443 & 0.348 & 0.390 & \\
OneR & Camera-3 & 0.545 & 0.599 & 0.571 & 0.6220\\
 & Camera-4 & 0.665 & 0.697 & 0.681 & \\
 & Camera-5 & 0.699 & 0.791 & 0.742 & \\ [1ex] 
\hline 
\end{tabular}}}}
\end{table}

\begin{table}[t!]


\centering
\caption{Experimental results using Feature Selection
\label{table:feature_selection_result}}{
\resizebox{\columnwidth}{!}{%
\begin{tabular}{c c c c c c} 
\hline\hline 
Algorithm & Camera & Precision & Recall & F-Measure & Overall Accuracy \\ [0.4ex] 
\hline 

 & Camera-1 & 1 & 1 & 1 &  \\      
 & Camera-2 & 1 & 0.999 & 1 & \\
Random Forest & Camera-3 & 1 & 1 & 1 & 0.9998\\
 & Camera-4 & 1 & 0.999 & 1 & \\
 & Camera-5 & 1 & 1 & 1 & \\ 
 & & & & & \\
  & Camera-1 & 0.997 & 0.997 & 0.997 &  \\   
 & Camera-2 & 0.992 & 0.991 & 0.997 & \\
J48 & Camera-3 & 0.997 & 0.997 & 0.998 & 0.9954\\
 & Camera-4 & 0.996 & 0.996 & 0.998 & \\
 & Camera-5 & 0.996 & 0.996 & 0.998 & \\ 
 & & & & & \\
 
  & Camera-1 & 0.882 & 0.922 & 0.9015 &  \\   
 & Camera-2 & 0.978 & 0.865 & 0.918 & \\
SVM & Camera-3 & 0.957 & 0.925 & 0.9407 &  0.8742\\
 & Camera-4 & 0.867 & 0.825 & 0.8454 & \\
 & Camera-5 & 0.823 & 0.927 & 0.8719 &\\
  & & & & & \\

  & Camera-1 & 0.808 & 0.928 & 0.864 &  \\   
 & Camera-2 & 0.827 & 0.621 & 0.710 & \\
Naive Bayes & Camera-3 & 0.763 & 0.874 & 0.815 & 0.8109\\
 & Camera-4 & 0.833 & 0.729 & 0.777 & \\
 & Camera-5 & 0.841 & 0.904 & 0.871 & \\ 
 & & & & & \\

  & Camera-1 & 0.722 & 0.693 & 0.707 &  \\   
 & Camera-2 & 0.443 & 0.348 & 0.390 & \\
OneR & Camera-3 & 0.545 & 0.599 & 0.571 & 0.6220\\
 & Camera-4 & 0.665 & 0.697 & 0.681 & \\
 & Camera-5 & 0.699 & 0.791 & 0.742 & \\ [1ex] 
\hline 
\end{tabular}}}

\end{table}

Next, we present our the performance evaluation metrics used in this work. We use \emph{Precision}, \emph{Recall} and \emph{F-Measure}, to evaluate the overall accuracy of source detection.
We assume the followings with respect to any source camera A. 
True Positives indicate frames of source A, detected to be of source A. 
True Negatives indicate correctly identified frames from sources other than A. 
False Positives indicate frames \emph{not} from source A, detected to be from source A. 
False Negatives indicate frames of source A, detected to be from some other source. 

We define Precision, Recall and F--Measure as the evaluation metrics, as follows:
\begin{equation}
\footnotesize Precision =\frac{True~Positive}{True~ Positive + False~ Positive}
\end{equation}
\begin{equation}
\footnotesize Recall =\frac{True~ Positive}{True~ Positive + False~ Negative} 
\end{equation}
\begin{equation}
\footnotesize F-Measure=\frac{2\times Precision \times Recall}{Precision + Recall}
\end{equation}

In our experiments, we use five different classifiers to evaluate video source detection accuracy, the comparative results of which are presented in Table~\ref{table:comparison}, in descending order of Detection Accuracy. 
Hence, in our work, the Random Forest classifier proves to achieve the best performance, closely followed by J48 classifier, as compared to the rest.

\begin{table}[t!]
\centering
\caption{Comparative Results for various Classifiers
\label{table:comparison}}{
\resizebox{\columnwidth}{!}{%
\begin{tabular}{c c c c} 
\hline\hline 
Algorithm & Using Feature Selection & Without Feature Selection & Overall Accuracy \\ [0.4ex] 
\hline 
Random Forest & yes & - & 99.98\%\\   
Random Forest & - & yes & 99.85\%\\
 & & & \\
J48 & yes & - & 99.54\%\\   
J48 & - & yes & 99.24\%\\
 & & & \\
SVM & yes & - & 87.42\%\\   
SVM & - & yes & 89.10\%\\
 & & & \\
Naive Bayes & - & yes & 84.49\%\\
Naive Bayes & yes & - & 81.09\%\\  
 & & & \\
OneR & yes & - & 62.20\%\\   
OneR & - & yes & 62.20\%\\ [1ex] 
\hline 
\end{tabular}}}
\end{table}

\begin{table}[]
\centering
\caption{Comparative results of proposed approach}
\label{tab:CompRes}
\scalebox{0.9}{
\begin{tabular}{|c|c|}
\hline
\textbf{Method}                          & \textbf{Classification Accuracy (\%)} \\ \hline
Goljan et.al.~\cite{goljan2016effect}      &         97.50 \\ \hline
     Chen et.al.~\cite{chen2015live}                &    98.25                                   \\ \hline
Proposed (for Random Forest classifier) & 99.98                                 \\ \hline
\end{tabular}}
\end{table}

We compare the performance of the proposed method, our experimental results, with other recent works of Goljan et. al.~\cite{goljan2016effect} and Chen et.al.~\cite{chen2015live}. 
The comparison results are presented in Table~\ref{tab:CompRes}, in terms of overall source classification accuracy, considering all five camera models.
It is evident from Table~\ref{tab:CompRes}, that the proposed feature based classification technique, performs video source identification with an accuracy considerably higher than the state--of--the--art.


\section{Conclusion and Future Work}
\label{sec:conclusion}

In this paper, we used machine learning classification techniques to authenticate and detect source of a digital video from a closed set of cameras. We compared the results from various classifiers. In our work, we stressed upon the development of an optimized model for video source classification. For testing and evaluation of the proposed model, we used a dataset of 5 cameras. An important outcome of this research, as evident from our experimental results, is that the best results are obtained from Random Forest and J48 classifiers.
Future research in this direction include experimenting with a larger dataset and also to perform source camera identification of videos which are downloaded from online social networks. 
\bibliographystyle{vancouver}
\bibliography{BibText}

\end{document}